\documentclass[11pt]{article}

\usepackage[final]{acl}
\usepackage{booktabs}   
\usepackage{multirow}   
\usepackage{graphicx}   
\usepackage{xcolor}     
\usepackage[table]{xcolor}

\usepackage{subfigure}
\usepackage{times}
\usepackage{latexsym}
\usepackage{tabularx}
\newcolumntype{Y}{>{\centering\arraybackslash}X}
\usepackage[T1]{fontenc}
\usepackage{amsmath,amssymb}  
\usepackage{algorithm}        
\usepackage{algpseudocode}    
\usepackage{dsfont}

\usepackage[utf8]{inputenc}

\usepackage{microtype}

\usepackage{inconsolata}
\usepackage{fontawesome5}
\usepackage{graphicx}
\usepackage{xcolor}
\definecolor{MorandiBlue}{RGB}{132, 171, 209}
\usepackage{xspace} 
\newcommand{\method}{\textsc{Lens}\xspace}
\usepackage{placeins}

\title{Less Noise, More Voice: Reinforcement Learning for Reasoning \\via Instruction Purification}

\usepackage{wasysym}
\usepackage{marvosym}
\usepackage{harmony}
\author{Yiju Guo$^{\eighthnote, \dagger}$, Tianyi Hu$^{\halfnote}$, Zexu Sun$^{\ViPa}$, Yankai Lin$^{\eighthnote}$\textsuperscript{\Letter}\\ 
  $^{\eighthnote}$ Gaoling School of Artificial Intelligence, Renmin University of China \\
  $^{\halfnote}$ Department of Computer Science, Aarhus University  \\
 $^{\ViPa}$  Baidu Inc. \\
    \Letter\texttt{\{yijuguo, yankailin\}@ruc.edu.cn} 
    \qquad \href{https://github.com/RUCBM/LENS}{\faGithub~\textbf{GitHub:} RUCBM/LENS}
    }
    
\begin{document}

\maketitle
\renewcommand{\thefootnote}{} 
\footnotemark\footnotetext{\Letter~Corresponding author: Yankai Lin.}
\footnotemark\footnotetext{$\dagger$~Work done during an internship at Baidu.}
\renewcommand{\thefootnote}{\arabic{footnote}} 

\begin{abstract}
Reinforcement Learning with Verifiable Rewards (RLVR) has advanced LLM reasoning, but remains constrained by inefficient exploration under limited rollout budgets, leading to low sampling success and unstable training in complex tasks. 
We find that many exploration failures arise not from problem difficulty, but from a small number of prompt tokens that introduce interference.
Building on this insight, we propose the \textbf{Less Noise Sampling Framework (\method)}, which first 
low-success prompts by identifying and removing interference tokens. then transfers successful rollouts from the purification process to supervise policy optimization on the original noisy prompts, enabling the model to learn to ignore interference in the real-world, noisy prompting settings. Experimental results show that \method significantly outperforms GRPO, delivering higher performance and faster convergence, with a 3.88\% average gain and over 1.6$\times$ speedup on math 
reasoning, and a 1.83\% gain on scientific and general reasoning. Our work highlights the critical role of pruning interference tokens in improving rollout efficiency, offering a new perspective for RLVR research.
\end{abstract}

\section{Introduction}
\label{sec: intro}

\begin{figure}[!ht]
    \centering
    \includegraphics[width=0.45\textwidth]{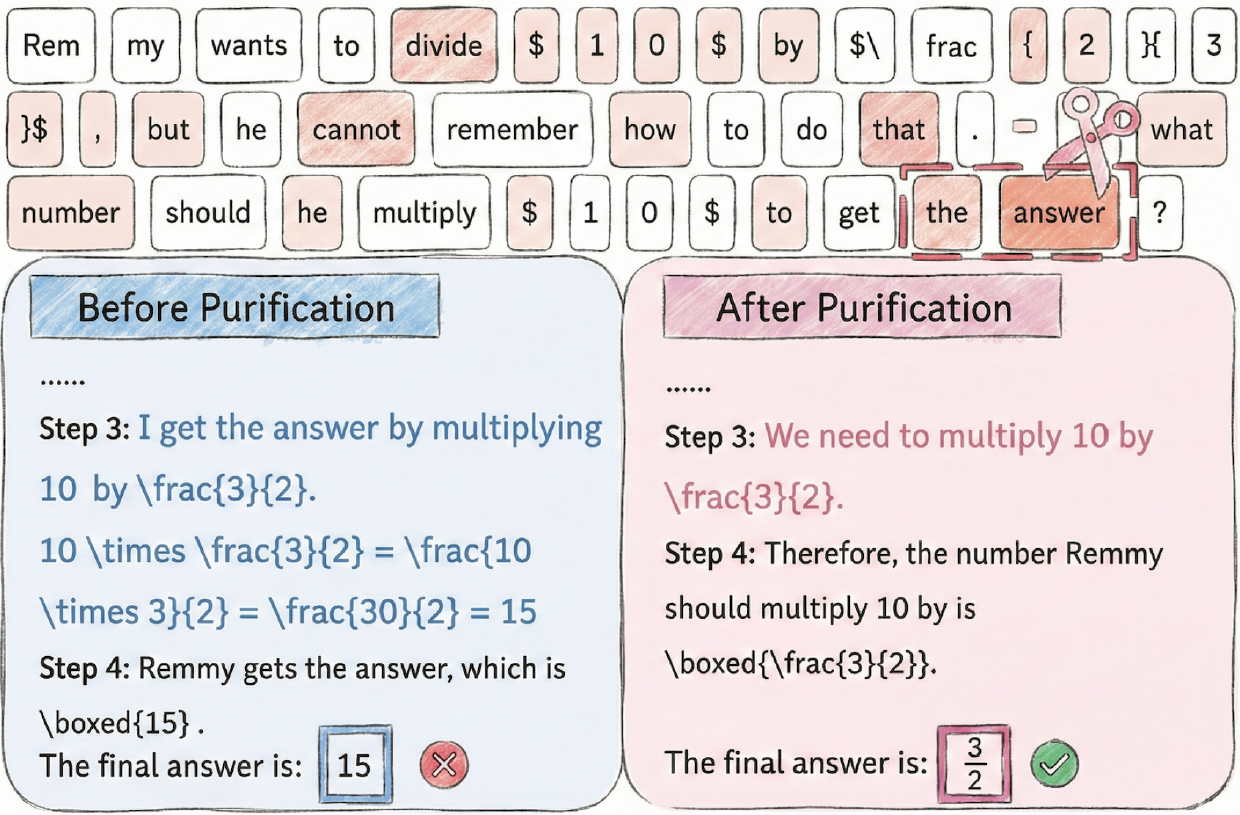} 
    \caption{An example of \textit{interference token purification}: Removing a few interference tokens corrects the reasoning rollout and turns it into a successful one.}
    \label{fig:examples}
    \vspace{-0.3cm}
\end{figure}
\begin{figure*}[!ht]
    \centering
    \includegraphics[width=1\textwidth]{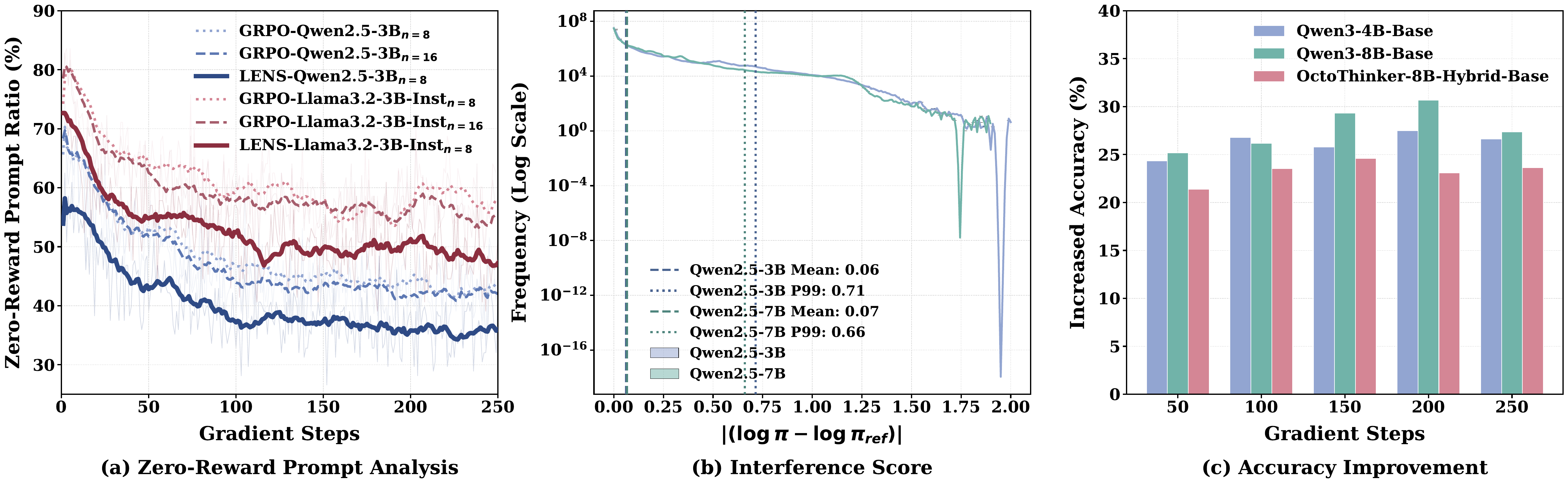} 
    \caption{\textbf{(a) Zero-Reward Prompt Analysis:} Comparison of the zero-reward prompt ratio across different models and rollout sizes ($n$). \method significantly reduces the proportion of zero-reward samples compared to GRPO, enhancing training efficiency.
\textbf{(b) Distribution of token-level Interference Scores} (log scale): Only a few tokens exhibit high interference. 
\textbf{(c) Rollout Accuracy Improvement:}  Removing these interference tokens leads to an improvement in rollout success rates (Average@8).}
    \label{fig:obervation}
    \vspace{-0.2cm}
\end{figure*}

Reinforcement Learning with Verifiable Rewards (RLVR), such as GRPO~\citep{shao2024deepseekmathrlvr}, has significantly advanced the reasoning capabilities of large language models (LLMs).
However, RLVR fundamentally relies on sampling correct rollouts to generate informative learning signals~\citep{yu2025daporlvr,greso}.
In complex reasoning tasks, reward sparsity arises from long-horizon decision making with delayed and binary feedback.
When combined with a high-dimensional action space, correct rollouts become exceedingly rare (Figure~\ref{fig:obervation}a), leading to a lack of positive samples and causing training to collapse or become highly inefficient~\cite{hare2019dealing}.

To mitigate these issues, recent works have primarily followed two directions: (1) scaling exploration by increasing rollout~\citep{xu2025notrolloutstrategy,yang2025depthrolloutstrategy,zhan2025exgrporolloutstrategy,xiong2025reinforcerolloutstrategy}, and (2) filtering zero-variance prompts~\citep{yu2025daporlvr,greso}.
However, the former has substantially higher computational cost without improving efficiency, while the latter sacrifices exploration on challenging samples, limiting the model’s ability to solve complex problems.
As a result, neither approach truly addresses the core issue of inefficient exploration on challenging samples.

To address this, we investigate why the model fails to explore successful rollouts. Through a fine-grained token-level analysis, surprisingly, we find that \textit{many failures arise not from problem difficulty, but from a few ($<5\%$) tokens that introduce excessive interference}, as shown in Figure~\ref{fig:examples} and Figure~\ref{fig:obervation}b. We define tokens that are likely to cause failures as \textit{Interference Tokens}. Simply pruning these tokens improves rollout accuracy on previously failed DeepMath~\citep{deepmath} samples by over 20\% across all the model families (Figure~\ref{fig:obervation}c).

Building on these insights, we introduce \textbf{\underline{Le}ss \underline{N}oise \underline{S}ampling Framework (\method)}, an online selective rollout framework that improves high-quality evolution by extracting informative learning signals from low-success prompts.  In the first stage, \method identifies and removes interference tokens within low-success prompts via interference score (Figure~\ref{fig:obervation}b), producing prompts that yield a higher proportion of successful rollouts.
In the second stage, \method performs a transfer from the purification process to the original noisy setting: successful rollouts generated under denoised prompts are used as high-reward supervision to calibrate policy optimization on the original prompts. Unlike standard filtering, this mechanism encourages the model to learn to ignore interference tokens, rather than merely fitting solutions under cleaner conditions, ultimately enhancing the robustness of LLM reasoning through self-exploration.

Experimental results show that \method significantly outperforms GRPO, achieving a \textit{Pareto improvement} in performance–efficiency trade-offs, with an average performance gain of 3.88\% and over 1.6$\times$ faster convergence across seven math reasoning benchmarks, with further gains on four out-of-domain general reasoning 
benchmarks. Furthermore, \method exhibits superior performance over both scaling exploration and prompt filtering baselines while using substantially fewer computational resources. These results further provide empirical support for our hypothesis: low-success, challenging prompts contain valuable training signals, highlighting the critical role of pruning interference tokens in improving rollout efficiency, offering a new perspective for RLVR research.

\begin{figure*}[!htbp]
    \centering
    \vspace{-0.3cm}
    \includegraphics[width=1\textwidth]{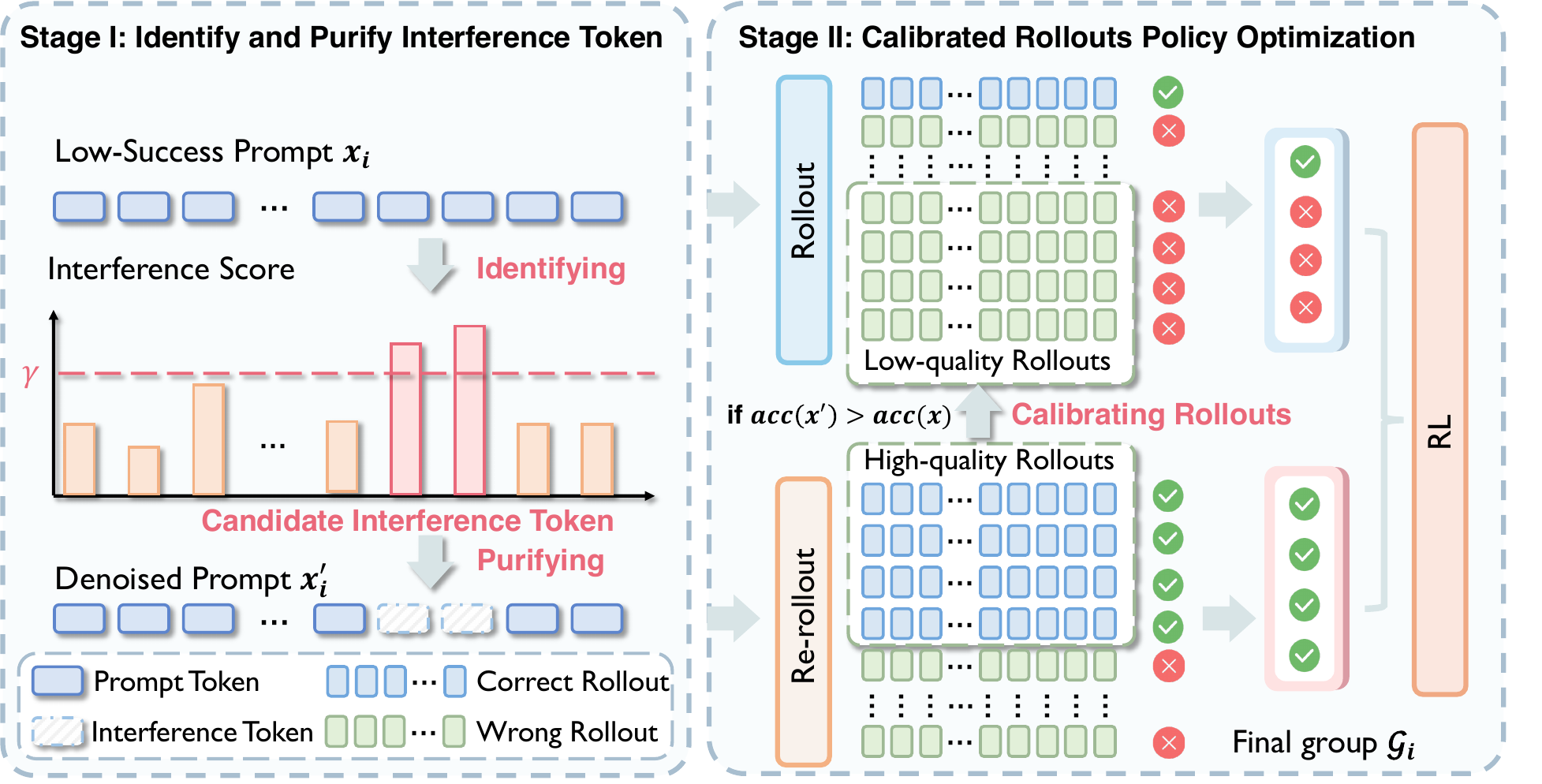} 
    \caption{\textbf{Method Overview.} In the first stage, \method identifies and purifies interference tokens within low-success prompts via \textit{Interference Score} (Defined in Section~\ref{sec:identification}), thereby generating a higher proportion of successful rollouts. In the second stage, \method uses successful rollouts from the denoised prompts as high-reward supervision to calibrate policy optimization on the original prompt, correcting gradient updates distorted by interference.}
    \vspace{-0.2cm}
    \label{fig:main}
    \vspace{-0.4cm}
\end{figure*}

\section{\method: Less Noise Sampling Framework }
\label{sec:method_objective}

We introduce \textbf{\method} (\underline{LE}ss \underline{N}oise \underline{S}ampling Framework), a plug-and-play rollout framework designed to facilitate effective policy exploration. 
\method consists of two key components: (1) \textit{Interference Token Identification and Purification}: identifying and purifying interference tokens in low-success prompts through interference score; (2) \textit{Calibrated Rollout Policy Optimization (CRPO)}, an efficient post-training algorithm that transfers learning signals from the denoised prompts produced by Component~(1) to the original noisy prompts, equipping the model with the ability to ignore interference and perform robust reasoning under noisy inputs.

\subsection{Interference Token Identification and Purification}
\label{sec:identification}

In this section, we first describe how interference tokens are identified and then explain how prompt purification is performed.

\paragraph{Interference Token Identification.}

We start from the observation that the reference model $\pi_{\text{ref}}$ provides a stable reference distribution learned from the training data.
In contrast, large token-level deviations of the learned policy $\pi_\theta$ from this reference often signal over-optimization or spurious behavior driven by noise, which can destabilize exploration~\cite{rafailov2024r}.

Motivated by this intuition, we identify interference tokens by measuring the token-level deviation between the current policy and the reference model.
Specifically, for a token prefix $s$ and the next generated token $a$, we define \textit{Interference Score} as:
\begin{equation}
\label{eq:interference_score}
S_I(s,a) \triangleq \bigl| \log \pi_\theta(a\mid s) - \log \pi_{\text{ref}}(a \mid s) \bigr|
\end{equation}

Tokens with large interference scores contribute disproportionately to the KL divergence from the reference distribution and are therefore treated as interference tokens.
Such deviations are commonly induced by reward over-optimization or noisy and misleading signals~\cite{rafailov2024r, gao2023scaling}, and can hinder effective exploration and generalization in the high-dimensional token action space~\cite{engstrom2020implementation, dai2025mitigating}.

\paragraph{Interference Purification.}
Based on the observations in Section~\ref{sec: intro} and the \textit{Interference Score}, we propose a token-wise inspection and pruning mechanism. By removing only a small number of tokens, this mechanism effectively eliminates interference while preserving the original semantics of the prompt with minimal impact.

Specifically, for the $i$-th prompt $x_i$, which contains $|x_i|$ tokens after tokenization, we compute token-level Interference Scores for all the tokens in the prompt and rank the tokens in descending order. Then, we introduce a deletion ratio $\gamma$ and select the top
$k=\lceil \gamma \cdot |x_i| \rceil$ tokens to form an interference token set $I_i$. We define the denoised prompt as $x'_i = x_i \setminus I_i$, which denotes removing all tokens in the interference set $I_i$. We set $\gamma$ to a small value (e.g., 1\%–5\%) to preserve the original semantics, with further discussion in Section~\ref{anaysis:threshold}.

\begin{algorithm*}[!ht]
\caption{CRPO: Calibrated Rollout Policy Optimization}
\label{alg:crpo}
\begin{algorithmic}[1]
\State \textbf{Input:} Policy $\pi_\theta$, reference policy $\pi_{\mathrm{ref}}$, dataset $\mathcal{D}$, group $m$, accuracy threshold $\tau$, pruning count $k$.
\For{iteration $t = 1, \dots, N$}
    \State Sample batch $\mathcal{B} = \{x_i\}_{i=1}^{B} \sim \mathcal{D}$.
    \For{each prompt $x_i \in \mathcal{B}$}
        \State Sample $m$ rollouts $Y_i = \{y_{i,1},\dots,y_{i,m}\} \sim \pi_\theta(\cdot \mid x_i)$.
        \State Partition $Y_i$ into success set $Y_i^+$ and failure set $Y_i^-$.
        \State Compute initial success rate $\bar{a}_i = |Y_i^+| / |Y_i|$. 
        \State Initialize training group $\mathcal{G}_i \leftarrow Y_i$ and prompt mapping $x^{\text{roll}}(y) \leftarrow x_i, \forall y \in \mathcal{G}_i$.
        
        \If{$\bar{a}_i < \tau$}
            \State \textbf{Interference Identification:} \Comment{\textcolor{MorandiBlue}{Interference Token Identification and Purification.} }
            \State Compute $S_I(t) = \lvert \log \pi_\theta(t \mid x_i) - \log \pi_{\mathrm{ref}}(t \mid x_i) \rvert$ for tokens in $x_i$ (Eq.~\ref{eq:interference_score}).
            \State Obtain $x'_i$ by deleting top-$k$ tokens in $x_i$ with highest $S_I(t)$.
            
            \State Sample $m$ rollouts $Y'_i \sim \pi_\theta(\cdot \mid x'_i)$ and compute success rate $\text{acc}(x'_i) = |Y'_i{}^+| / m$.
            
            \If{$\text{acc}(x'_i) > \bar{a}_i$}
                \State Let $P_i$ be successful rollouts in $Y'_i$.
                \State Select $R_i \subseteq Y_i^-$ with $|R_i| = \min(|Y_i^-|, |P_i|)$.
                \State Update $\mathcal{G}_i \leftarrow (Y_i \setminus R_i) \cup P_i$.
                \State Update $x^{\text{roll}}(y) \leftarrow x'_i$ for all $y \in P_i$.
            \EndIf
        \EndIf
        
        \State \textbf{Policy Calibration:}\Comment{\textcolor{MorandiBlue}{Calibrated Rollout Policy Optimization.} }
        \State Compute importance ratios $\rho(y;\theta)$ (Eq.~\ref{eq:important_sampling}) and weights $\tilde{w}(y)$ based on $\bar{a}_i$ (Eq.~\ref{eq:weight}).
        \State Compute calibrated advantages $\hat A(y)$ (Eq.~\ref{eq:advantage}) and update $\pi_\theta$ using $\mathcal{L}(\theta)$ (Eq.~\ref{eq:crpo}).
    \EndFor
\EndFor
\end{algorithmic}
\end{algorithm*}
\vspace{-0.3cm}

\subsection{Calibrated Rollout Policy Optimization}
\label{sec:final_obj}

While interference tokens can be identified, directly removing them is not always beneficial, as only about 20\% of prompts exhibit improvement in rollout accuracy after removal (Figure~\ref{fig:obervation}c). 

Therefore, we adopt Calibrated Rollout Policy Optimization (\textbf{CRPO}),  which applies interference token purification (Section~\ref{sec:identification}) to obtain rollouts with a higher proportion of successful samples.
When the original prompt exhibits low sampling success (i.e., success rate below $\tau$; see Appendix~\ref{apex: tau} for sensitivity analysis), CRPO treats rollouts generated from the denoised prompt $x'_i$ as a source of transferable supervision and applies this signal to guide policy optimization on the original prompt $x_i$, enabling interference-aware calibration.
Such calibration equips the model to recognize interference under noisy prompts and thereby prevents training collapse.
The complete algorithmic procedure is provided in Algorithm~\ref{alg:crpo}.

\paragraph{Sample Reweighting.}
To mitigate the impact of interference-induced failures and properly incorporate the successful rollouts of the denoised prompt, we adopt a sample reweighting strategy to calibrate the training signal.

Let $Y_i = Y_i^+ \cup Y_i^-$ denote the set of rollouts sampled from the original prompt $x_i$,
where $Y_i^+$ and $Y_i^-$ are the sets of successful and failed rollouts, respectively.
We define the initial sampling success rate as
$\bar{a}_i = |Y_i^+| / |Y_i|$.

In addition, we collect a set of successful rollouts $P_i$ by sampling from the denoised prompt $x'_i$.
To ensure that pruning provides a genuine improvement, we activate rollout replacement only when
the denoised prompt achieves higher empirical accuracy:
\begin{equation}
g_i = \mathbb{I}\big[\mathrm{acc}(x'_i) > \mathrm{acc}(x_i)\big].
\end{equation}
When $g_i = 1$, we replace a subset of failed rollouts in $Y_i^-$ with successful ones from $P_i$.
Specifically, we randomly sample a subset $R_i \subseteq Y_i^-$ with
$|R_i| = \min(|Y_i^-|, |P_i|)$.
Let $\mathcal{G}_i$ be the reconstructed rollout set:
\begin{equation}
\mathcal{G}_i = 
\begin{cases} 
Y_i^+ \cup (Y_i^- \setminus R_i) \cup P_i, & \text{if } g_i = 1, \\
Y_i, & \text{otherwise.}
\end{cases}
\end{equation}
We define the unnormalized weight $\tilde{w}(y)$ for each rollout $y \in \mathcal{G}_i$ by applying $\bar{a}_i$ as a scaling factor:
\begin{equation}
\label{eq:weight}
\tilde{w}(y) = 
\begin{cases} 
\bar{a}_i, & y \in Y_i^+, \\[6pt]
1 - \bar{a}_i, & y \in P_i \cup (Y_i^- \setminus R_i).
\end{cases}
\end{equation}

\paragraph{Objective Function.}
To enable transfer of learning signals from the denoised prompt $x'_i$ back to the original prompt $x_i$, we formulate a unified objective that corrects for distribution mismatch and stabilizes policy optimization. Let $x^{\text{roll}}(y) \in \{x_i, x'_i\}$ denote the prompt variant used to sample rollout $y$
under the rollout policy $\pi_{\text{old}}$.
We apply importance correction by defining the ratio:
\begin{equation}
\label{eq:important_sampling}
\rho(y;\theta) = \frac{\pi_\theta(y \mid x_i)}{\tilde w(y)\pi_{\text{old}}(y \mid x^{\text{roll}}(y))}.
\end{equation}

We compute group-relative advantages by normalizing rewards over the reconstructed rollout set $\mathcal{G}_i$:
\begin{equation}
\label{eq:advantage}
\hat{A}(y) = \frac{r(y) - \text{mean}_{y' \in \mathcal{G}_i}\bigl[r(y')\bigr]}{\text{std}_{y' \in \mathcal{G}_i}\bigl[r(y')\bigr]}.
\end{equation}

Finally, we optimize a PPO-style clipped surrogate objective with KL regularization:
\begin{equation}
\label{eq:crpo}
\begin{aligned}
\mathcal{L}(\theta) =
&-\sum_{y \in \mathcal{G}_i}
\min\Big(
\rho(y;\theta)\hat A(y),\\
&\quad\textsc{clip}(\rho(y;\theta), 1-\epsilon, 1+\epsilon)\hat A(y)
\Big)\\
&+\beta\,\mathbb{D}_{\mathrm{KL}}\!\left(
\pi_\theta(\cdot \mid x_i)\,\|\,\pi_{\mathrm{ref}}(\cdot \mid x_i)
\right).
\end{aligned}
\end{equation}

Overall, CRPO provides a selective and stable calibration mechanism that improves policy optimization by exploiting high-quality rollouts revealed through interference purification.

\section{Experiments}
\setlength\tabcolsep{3.5pt}
\begin{table*}[!t]
    \centering
    {
    \begin{tabularx}{\textwidth}{l YYYYYYYY}
        \toprule
        \multicolumn{1}{c}{\multirow{2}{*}{\textbf{Model}}} & \multicolumn{4}{c}{\textbf{Pass@1}} & \multicolumn{3}{c}{\textbf{Average@16}} & \multirow{2}{*}{\textbf{Avg.}} \\
        \cmidrule(lr){2-5} \cmidrule(lr){6-8}
        & \textbf{MATH} & \textbf{Minerva} & \textbf{Olympiad} & \textbf{GAO} & \textbf{AMC23} & \textbf{AIME24} & \textbf{AIME25} & \\
        \midrule
        \textbf{Llama3.2-3B-Instruct} & 40.02 & 16.54 & 15.26 & 35.58 &10.16 &1.88 & 0.00  & 17.06 \\
        + GRPO & 51.60 & \underline{21.69} & 20.00 & 44.68 & 22.81& 6.25 & \underline{0.83} & 23.98 \\
        + DAPO & 53.00 & \underline{21.69} & 19.26 & 47.01 & 26.09 &9.79 &0.42 & 25.32\\
        + GRESO &51.80  & \underline{21.69} & 19.70 & 45.45 & 26.09& 6.46&\textbf{1.04} &24.60 \\
        + GRPO$_{\text{extended}}$ & 51.20 & 20.96 & 19.41 & 44.68 & 26.76& 6.25 & \textbf{1.04} &  24.33 \\
        + DAPO$_{\text{extended}}$ & 52.20 & 20.22 &\underline{20.15} &\underline{48.31}  & \underline{27.34}&\underline{10.21}& 0.42& \underline{25.55} \\
        + GRESO$_{\text{extended}}$ &\underline{53.20}  & 18.38 &18.81 & 46.75 &27.19 &7.29 & \underline{0.83}& 24.64\\
        \rowcolor{gray!10}+ \textsc{Lens}$_{\text{GRPO}}$ & \textbf{55.80} & \textbf{22.43} & \textbf{21.04} & \textbf{48.83} & \textbf{29.84} & \textbf{10.62} & \underline{0.83}  & \textbf{27.03}\\
        \midrule
        \textbf{Qwen3-4B-Base} & 72.20 & 27.21 & 38.96 & 65.19 & 26.72& 5.62&6.70 &34.66 \\
        + GRPO & 79.40 & 34.56 & 45.19 & \underline{70.91} & 53.28&12.71 &\underline{14.79} &44.41 \\
        + DAPO & 80.20 & 35.29 & 42.96 & 68.57 & 52.97& 13.12&11.04 & 43.50\\
        + GRESO & \underline{80.80} & 36.76 & \underline{45.33} & 69.61 &55.78 &16.46 & 12.50& 45.32\\
        + GRPO$_{\text{extended}}$ & 78.20 & \underline{39.34}&44.30 & 70.39 &54.53 &12.92 &11.88 &44.51 \\
        + DAPO$_{\text{extended}}$ &\underline{80.80}  & \underline{39.34} & \underline{45.33}& 68.57 & 53.44& 16.04& 11.04&44.94 \\
        + GRESO$_{\text{extended}}$ &80.60  & 37.50 &44.74 & 69.06 &\underline{58.75} & \underline{18.33}& 12.71&\underline{45.96} \\
        \rowcolor{gray!10}+ \textsc{Lens}$_{\text{GRPO}}$ & \textbf{83.20} & \textbf{41.54}&\textbf{50.37} &\textbf{72.99}&\textbf{60.94}&\textbf{18.54}&\textbf{16.46} &\textbf{49.15} \\
        \midrule
        \textbf{Qwen3-8B-Base} & 77.00 & 34.29 &41.19 & 64.68 & 38.59&5.62& 5.83& 38.17 \\
        + GRPO & 84.00 & \underline{45.22} &48.15 & 73.25 & 58.91& 18.33& 18.75&49.52 \\
        + DAPO &85.00  & 44.12 &52.00 & 73.51 & 62.03& 21.46& 17.92&50.86 \\
        + GRESO &84.20  &  41.18&49.93 & 71.69 & 64.69&20.83&15.42 & 49.71\\
        + GRPO$_{\text{extended}}$ & \underline{85.20} & 43.38 & 52.15& \textbf{76.01} &\underline{66.25} & 22.08& 19.17& \underline{52.03}\\
        + DAPO$_{\text{extended}}$ &85.00  & 44.49 &\underline{52.44} & 74.55 &65.47&22.50 &18.75  & 51.88\\
        + GRESO$_{\text{extended}}$ 
        & 84.60 & 43.38 & 50.81& \underline{75.32} &60.62 & 21.46&\underline{19.38} &50.80 \\
        \rowcolor{gray!10}+ \textsc{Lens}$_{\text{GRPO}}$ & \textbf{86.00} & \textbf{48.16} &\textbf{53.33} & \underline{75.32} &\textbf{66.56}&\textbf{23.96} &\textbf{20.21} & \textbf{53.36}\\
        \bottomrule
    \end{tabularx}}
    \caption{\textbf{Detailed evaluation results on seven math reasoning benchmarks.} The best and second best results are in \textbf{bold} and \underline{underlined}. (*) Even in the unfavorable setting where GRPO$_{\text{extended}}$, DAPO$_{\text{extended}}$ and GRESO$_{\text{extended}}$ are trained for $2\times$ more rollouts, \method still outperforms them on the majority of benchmarks.}
    \label{tab:main}
\end{table*}
\begin{table*}[!t]
    \centering
    \begin{tabularx}{\textwidth}{l YYYYY}
        \toprule
        \textbf{Model} & \textbf{SuperGPQA} & \textbf{GPQA-D} & \textbf{BBEH} & \textbf{MMLU-Pro} & \textbf{Avg.} \\
        \midrule
        \textbf{Llama3.2-3B-Instruct} & 18.49 & 25.25 & 5.53 & 34.69 & 20.99 \\
        + GRPO & \underline{18.77} & \underline{29.29} & \underline{6.99} & \underline{36.84} & \underline{22.97} \\
        \rowcolor{gray!10}+ \textsc{Lens}$_{\text{GRPO}}$ & \textbf{20.05} & \textbf{30.17} & \textbf{8.34} & \textbf{38.21} & \textbf{24.19} \\
        \midrule
        \textbf{Qwen3-4B-Base} & 25.99 & 26.30 & 8.05 & 51.60 & 27.99 \\
        + GRPO & \underline{30.17} & \underline{40.40} & \underline{10.88} & \underline{59.35} & \underline{35.20} \\
        \rowcolor{gray!10}+ \textsc{Lens}$_{\text{GRPO}}$ & \textbf{31.66} & \textbf{43.94} & \textbf{11.95} & \textbf{60.65} & \textbf{37.05} \\
        \midrule
        \textbf{Qwen3-8B-Base} & 30.40 & 33.30 & 10.55 & 58.00 & 33.06 \\
        + GRPO & \underline{35.26} & \underline{48.99} & \underline{13.08} & \underline{62.41} & \underline{39.94} \\
        \rowcolor{gray!10}+ \textsc{Lens}$_{\text{GRPO}}$ & \textbf{35.59} & \textbf{55.56} & \textbf{13.36} & \textbf{64.88} & \textbf{42.35} \\
        \bottomrule
    \end{tabularx}
    \caption{\textbf{Out-of-domain generalization results.} Models are trained only on mathematical data and evaluated on scientific reasoning (SuperGPQA, GPQA-D) and general complex reasoning (BBEH, MMLU-Pro). GPQA-D denotes GPQA-Diamond. The best and second best results are in \textbf{bold} and \underline{underlined}.}
    \label{tab:generalization}
\end{table*}
\begin{figure*}[!ht]
    \centering
    \includegraphics[width=1\textwidth]{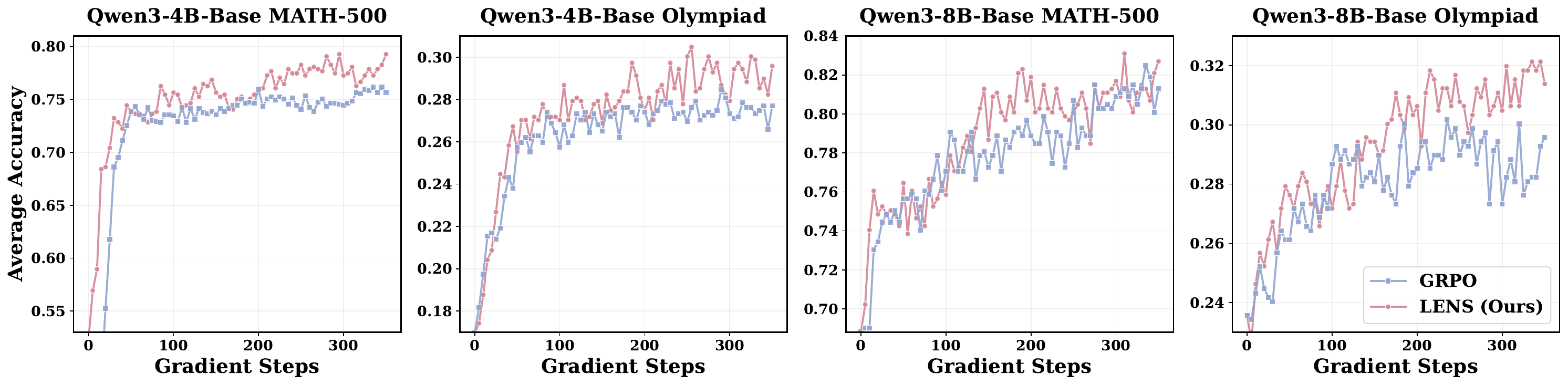} 
    \caption{\textbf{Learning curves of \textsc{Lens} and GRPO across model scales and task difficulties.} We compare Qwen3-4B/8B-Base backbones on MATH-500 (Medium) and OlympiadBench (High). \textsc{Lens} converges faster and achieves comparable or higher final accuracy than GRPO under the same training step, indicating more efficient optimization.}
    \label{fig:result}
\end{figure*}

\subsection{Experiment Settings}
\noindent\textbf{Base Models.} 
We conduct experiments across three distinct model families, Llama-3.2~\cite{meta33llama}, Qwen-2.5~\cite{team2024qwen2} and Qwen-3~\cite{yang2025qwen3}, covering various parameter scales. Specifically, we utilize five models: Llama-3.2-3B-Instruct~\cite{meta33llama}, Qwen2.5-3B, Qwen2.5-7B~\cite{team2024qwen2}, Qwen3-4B-Base and Qwen3-8B-Base~\cite{yang2025qwen3}. 

\paragraph{Baseline.} 
We evaluate \method against vanilla GRPO \cite{shao2024deepseekmathrlvr} and two representative strategies for handling zero-variance prompts: increased sampling and zero-variance filtering. For increased sampling, we implement GRPO$_{\text{extended}}$, which doubles the rollout budget per instruction. For zero-variance filtering, we compare with DAPO \cite{yu2025daporlvr} and GRESO \cite{greso}, maintain training stability by discarding or skipping zero-variance prompts via post-rollout filtering and pre-rollout prediction, respectively. We also report their extended variants, DAPO$_{\text{extended}}$ and GRESO$_{\text{extended}}$, which run twice the number of training epochs.
Notably, \method operates under a strictly lower computational budget, without increasing rollout counts or training epochs.

\paragraph{Training and Evaluation Datasets.} 
For the RL training phase, we employ Openr1-Math-46k~\cite{luffy}, a large-scale, high-quality dataset designed for mathematical reasoning. In the evaluation, we assess model performance across a diverse set of seven benchmarks: MATH500~\cite{hendrycks2021measuring}, AMC23~\cite{amc2023}, AIME24, AIME25~\cite{li2024numinamathaime}, GaokaoEN-2023~\cite{zhang2023evaluatinggaokao}, Minerva~\cite{lewkowycz2022solvingMineva}, and OlympiadBench~\cite{he2024olympiadbench}. This selection covers a broad range of difficulty levels, enabling comprehensive evaluation. To further assess out-of-domain generalization, we additionally evaluate on scientific reasoning (SuperGPQA~\cite{supergpqa}, GPQA-Diamond~\cite{gpqa}) and general complex reasoning benchmarks (BBEH~\cite{bbeh}, MMLU-Pro~\cite{mmlupro}).

\paragraph{Training Settings.} 
We leverage GRPO as the basis for \method, setting the KL coefficient $\beta=0.001$ and the imitation coefficient $\gamma=0.001$. Both preliminary and validation experiments are conducted using Llama-3.2, Qwen-2.5, and Qwen-3 models. We configure the maximum response length to 4096 tokens, the learning rate to $1 \times 10^{-6}$, both the rollout and update batch sizes to $128$, the number of rollouts to 8, top-$p$ to 1, and the temperature to 1. Detailed training hyperparameters for GRPO are provided in Appendix~\ref{apex:appendix_training_settings}.

\paragraph{Evaluation Settings.} 
For evaluation, we set both the temperature and top-$p$ to $1.0$, with a maximum generation length of $4,096$ tokens. We primarily report the \texttt{Pass@1} accuracy. To ensure robust evaluation on high-difficulty benchmarks such as AMC23, AIME24, and AIME25, we present the results averaged over 16 generation samples.

\subsection{Main Results and Analysis}
Table~\ref{tab:main} and Table~\ref{tab:generalization} summarize the performance on Qwen3-4B-Base, Qwen3-8B-Base and Llama-3.2-3B-Instruct, while results on
Qwen2.5-3B and Qwen2.5-7B are included in Appendix~\ref{tab:qwen2.5}.
We draw two conclusions:
(1) \textbf{Superiority over rollout-intensive baselines.} \method consistently outperforms GRPO~\citep{shao2024deepseekmathrlvr} and GRPO$_{\text{extended}}$ given the same training corpus. 
This indicates that simply increasing the rollout budget is insufficient for generating informative samples when exploration is affected by interference tokens. 
In contrast, \method improves sample efficiency by enabling the model to focus on critical information, thereby effectively enhancing its reasoning performance.
(2) \textbf{Advantage over filtering strategies.} \method also surpasses the pre-rollout filter GRESO~\citep{greso} and the post-rollout filter DAPO~\citep{yu2025daporlvr}, including in settings with fewer rollouts.
These results suggest that aggressively discarding zero-variance prompts can limit capability expansion, particularly on challenging benchmarks, which explains why \method yields larger improvements on such datasets (e.g. AMC23 \& AIME24).
(3) \textbf{Effectiveness on general reasoning tasks.} \method consistently outperforms GRPO on both Qwen3 and Llama3.2 families across SuperGPQA, GPQA-Diamond, BBEH, and MMLU-Pro, with an average gain of +1.83\%. This demonstrates that interference token purification improves general reasoning capability rather than overfitting to math-specific patterns.

Figure~\ref{fig:result} illustrates the learning curves for Qwen3-4B-Base and Qwen3-8B-Base across MATH-500 and OlympiadBench. \method consistently achieves higher progressive and final accuracy compared to GRPO across all configurations. Notably, on the more challenging OlympiadBench, \method exhibits more stable and continuous improvement, in contrast to the fluctuations observed with GRPO. We attribute this advantage to two primary factors:
(1) \method enhances the model's exploration capacity by effectively removing potential interference tokens, leading to improved sample quality, and facilitating more effective exploration of capability boundaries;
(2) By contrasting correct responses from interference-free rollouts with erroneous responses in the original sampling, \method helps the model focus on key information, thereby boosting its reasoning abilities.

\section{Further Analysis}

\begin{figure*}[!ht]
    \centering
    \includegraphics[width=1\textwidth]{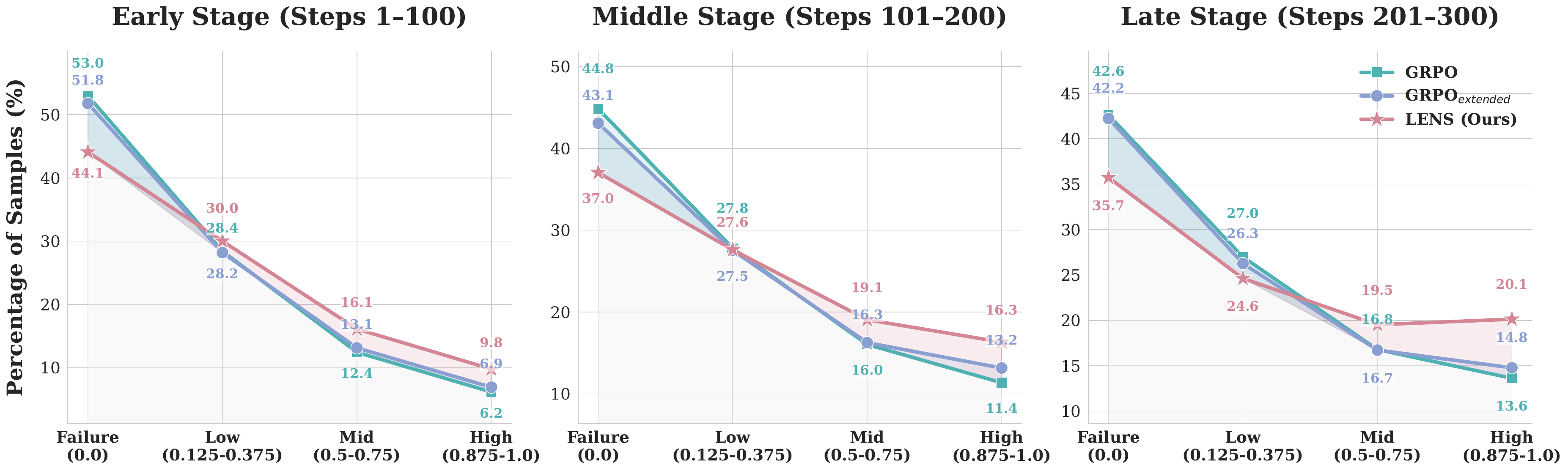} 
    \caption{\textbf{Sampling accuracy distribution across three training phases.} We compare the sampling distributions of GRPO, GRPO$_\text{extended}$ and \method across the early, middle and late training stages.}
    \label{fig:evolution}
\end{figure*}
\begin{figure*}[!ht]
    \centering
    \includegraphics[width=1\textwidth]{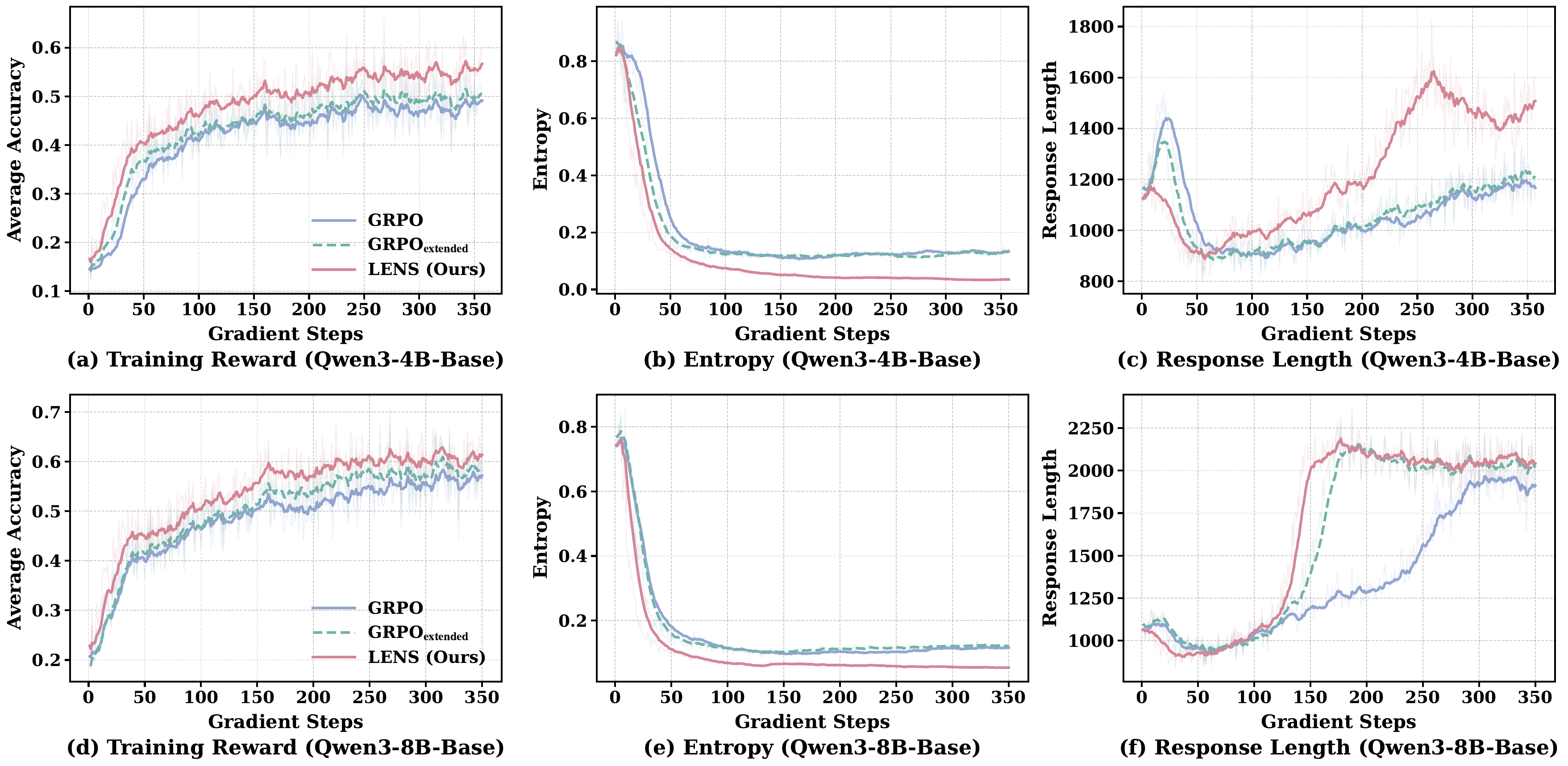} 
    \caption{\textbf{Training dynamics across different model scales.} Each row reports average accuracy, entropy, and response length during training for Qwen3-4B-Base (top) and Qwen3-8B-Base (bottom). Compared with GRPO and GRPO$_\text{extended}$, \method exhibits more consistent and stable trends.}
    \label{fig:training_dynamic}
\end{figure*}
\begin{figure*}[!ht]
    \centering
    \includegraphics[width=1\textwidth]{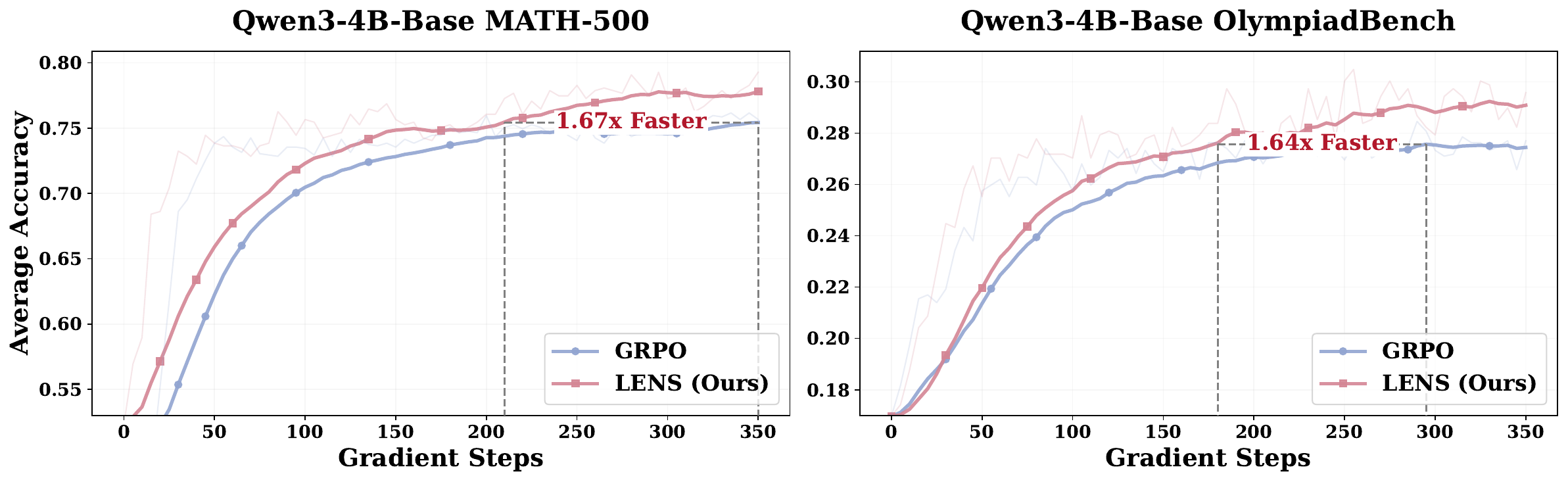} 
    \caption{\textbf{Training efficiency comparison of \method and GRPO on MATH-500 and OlympiadBench.} The gray dashed lines indicate the number of training steps required for both methods to reach the highest average accuracy of the
baseline during the entire training period. \method demonstrates superior sample efficiency and faster convergence.}
    \label{fig:efficient_4b}
    \vspace{-0.3cm}
\end{figure*}

In this section, we conduct training dynamics analysis (Section~\ref{anaysis:training_dynamics}), efficiency analysis (Section~\ref{anaysis:efficitency}), threshold sensitivity analysis (Section~\ref{anaysis:threshold}), pruning strategies analysis (Appendix~\ref{anaysis:masking}) and computational overhead (Appendix~\ref{anaysis:computation}).

\subsection{Training Dynamics Analysis}
\label{anaysis:training_dynamics}
We analyze the training dynamics of \method in two complementary views: (1) the sampling accuracy distribution across training stages (Figure~\ref{fig:evolution}), and (2) the evolution of training reward, policy entropy, and response length (Figure~\ref{fig:training_dynamic}). Together, they characterize both the \emph{quality} of rollouts and the \emph{trajectory} of policy learning under \method.

\paragraph{Higher-quality rollouts throughout training.} 
We first examine the sampling accuracy distribution across three training stages: early (Steps 1--100), middle (Steps 101--200), and late (Steps 201--300). As shown in Figure~\ref{fig:evolution}, \method substantially reduces the proportion of zero-reward prompts (\emph{Failure} category) compared to GRPO and GRPO$_\text{extended}$ across all stages. Correspondingly, more prompts are shifted into the \emph{Mid} and \emph{High} categories, which contain more informative learning signals.

\paragraph{More stable and decisive policy learning.} 
Beyond rollout quality, \method also exhibits healthier learning trajectories. (1) \textbf{Stable accuracy progression.} In Figures~\ref{fig:training_dynamic}(a, d), \method achieves steadier accuracy improvements than GRPO and GRPO$_\text{extended}$ across model scales. 
(2) \textbf{Emergence of confident long-form reasoning.} \method yields more rapid gains in both accuracy and response length while maintaining moderate entropy~\citep{cheng2025reasoning} (Figure~\ref{fig:training_dynamic}), with the earlier emergence of long-form reasoning (``aha moment''~\citep{guo2025deepseekrlvr}) reflecting more decisive reasoning.

\subsection{Efficiency Analysis}
\label{anaysis:efficitency}
\begin{figure*}[!ht]
    \centering
    \includegraphics[width=1\textwidth]{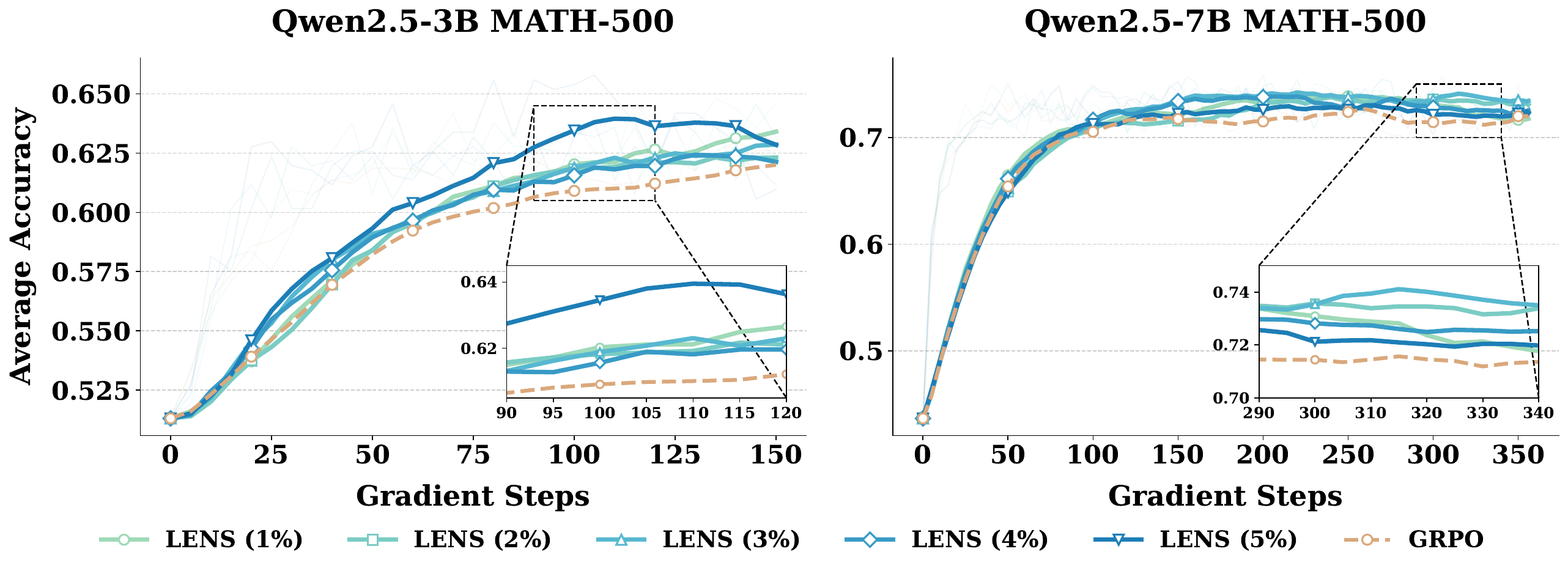} 
    \caption{\textbf{Performance convergence of \method on MATH-500 with Qwen2.5-3B/7B.} \method matches or exceeds GRPO throughout training. Insets highlight that the optimal threshold depends on model capacity: the weaker 3B model requires a higher threshold, while the stronger 7B model achieves optimal results with a lower threshold.}
    \label{fig:threshold}
    \vspace{-0.3cm}
\end{figure*}
We evaluate the learning efficiency of \method relative to GRPO on MATH-500 and OlympiadBench. We quantify efficiency by the relative speedup in terms of gradient steps required to reach the peak average accuracy attained by the GRPO baseline.

\paragraph{Results.} 
Figure~\ref{fig:efficient_4b} illustrates the comparative learning curves of \method and GRPO. 
Notably, \method achieves the same peak performance as GRPO while requiring $1.67\times$ fewer gradient steps on MATH-500 and attaining a $1.64\times$ acceleration on OlympiadBench. 
We attribute this substantial gain in training efficiency to the enhanced quality of the rollouts generated during exploration. 
By effectively mitigating noise from interference tokens, \method provides more stable and informative rollouts, which significantly accelerates convergence compared to the vanilla GRPO.

\subsection{Threshold Sensitivity Analysis}
\label{anaysis:threshold}

We conduct a sensitivity analysis on the pruning threshold for interference tokens using Qwen2.5-3B and Qwen2.5-7B on MATH-500. Figure~\ref{fig:threshold} shows validation accuracy curves for various thresholds ranging from 1\% to 5\%, alongside the vanilla GRPO baseline. Detailed sensitivity analysis of the success rate threshold $\tau$ in Appendix~\ref{apex: tau}.

\paragraph{Results.} 
We observe two key findings:
(1) \method consistently matches or outperforms the GRPO throughout the entire training process across all evaluated thresholds (1\%--5\%). 
(2) As highlighted in the insets, the optimal threshold exhibits a clear correlation with model capacity: the Qwen2.5-3B model yields superior performance with a higher threshold, whereas the more capable Qwen2.5-7B model achieves optimal performance with a lower threshold. 
This pattern suggests that models with limited capacity are more strongly affected by interference tokens.

\section{Related Work}

\noindent\textbf{Credit Assignment.} 
Credit assignment~\citep{kazemnejad2024vineppocredit, bentegeac2025tokenlogprob,chai2024macredit} is crucial for improving Reinforcement Learning with Verifiable Rewards (RLVR)~\citep{guo2025deepseekrlvr,liu2025understandingrlvr,yue2025vaporlvr,yu2025daporlvr,shao2024deepseekmathrlvr,wen2025reinforcementrlvr}, helping the model identify key decision paths that contribute most to the final decision. 
Credit assignment methods use forward signals, such as attention~\citep{chan2024dense, li2025attention}, log probabilities~\citep{bentegeac2025tokenlogprob}, and entropy~\cite{tan2025gtpoentropy}, to evaluate the importance of each token based on the model's forward pass.
However, existing methods primarily focus on output tokens during generation, while overlooking how distracting information within instruction tokens can mislead model behavior~\cite{guo2025learning}.
To address this gap, we propose extending the log-probability signal to the instruction level, analyzing and removing low-information instruction tokens to enhance the model’s focus during inference. This approach effectively improves the model’s sampling success rate and sample diversity.

\noindent\textbf{GRPO Signal Collapse.} 
GRPO suffers from signal collapse that identical rewards in a group lead to vanishing gradients, effectively halting further learning. 
Current studies can be categorized into 4 directions: 
(1) \textbf{Prompt Filtering}, removing extremely difficult prompts to maintain a productive training set;~\citep{yu2025daporlvr, xiong2025minimalistpromptfilter, zheng2025actpromptfilter};
(2) \textbf{Scaling Exploration}, dynamically allocating a greater number of rollouts to harder samples, thereby increasing reward variance within groups~\citep{xu2025notrolloutstrategy,yang2025depthrolloutstrategy,zhan2025exgrporolloutstrategy,xiong2025reinforcerolloutstrategy}; and 
(3) \textbf{Reward Function Design}, revising the advantage computation to prevent vanishing gradients~\cite{le2025noadvantage}.
These approaches focus on optimizing rollout strategies or reward functions while overlooking the interference  tokens in the instruction that degrade sample quality.
To address this gap, we propose a sampling framework to dynamically identify and purify these tokens, mitigating signal collapse and improving training efficiency.

\vspace{-0.1cm}
\section{Conclusion}
\vspace{-0.1cm}
In this paper, we first reveal a novel observation: in many cases, RLVR is only a few interference tokens away from discovering correct rollouts.
Building on this insight, we propose \method to identify interference tokens and to transfer successful rollouts generated from denoised prompts to calibrate policy optimization on the original noisy prompts.
Extensive experiments demonstrate that \method consistently outperforms GRPO in both performance and efficiency, with a 3.88\% average gain and over 1.6$\times$ speedup. 
\method also exhibits better performance over both scaling exploration and prompt filtering baselines, and generalizes to scientific and general reasoning tasks with a 1.83\% average gain.
Overall, our findings offer a fundamentally new perspective on improving sampling efficiency in RLVR and open up promising directions for future research.

\section*{Limitations}
Our work has the following limitations:
(1) \textbf{Model Scale Constraints:} Due to limited computational resources, our experiments were conducted on models with up to 8B parameters. Evaluating the performance and scalability of our method on larger-scale models (e.g., 32B or 70B) remains an avenue for future research.
(2) \textbf{Binary Reward Limitation:} The effectiveness of our approach has been validated primarily in tasks with binary rewards. Its applicability to more complex environments, such as those with multi-dimensional scoring, requires further investigation.
(3) \textbf{Algorithmic Integration:} While we demonstrated the efficacy of our method within the GRPO framework, we have not yet explored its integration with other GRPO-based variants. Specifically, our method could be combined with algorithms that optimize rollout frequency or reward functions, potentially enhancing exploration capabilities and training stability.

\section*{Acknowledgments}
This work was supported by The National Natural Science Foundation of China (No.\ 62376273 and No.\ U2436209).
The research is also funded by the Pioneer Centre for AI and the Novo Nordisk Foundation (NNF24OC0099109). We also thank the Department of Computer Science at Aarhus University for providing Tianyi Hu's travel support.

We sincerely thank all the anonymous reviewers and (S)ACs for their constructive comments, Wentong Chen, Yupeng Huo and Bokai Ji for their help with the implementation, and Tingchen Fu for valuable feedback on the writing.


\bibliography{custom}
\newpage
\appendix

\section{Detailed Training Settings}
\label{apex:appendix_training_settings}
The complete training hyper-parameters in GRPO and \method are put in Table~\ref{tab: basic hyper-parameters}.

\begin{table}[!htp]  
\begin{center}
\begin{tabular}{ll}
\toprule
Hyper-parameter & Value \\
\midrule
Train Batch Size & 128  \\
Micro Batch Size & 128 \\
Rollout $n$ & 8 \\
Maximum Prompt Length & 512 \\
Maximum Response Length & 4096 \\
Temperature & 1.0 \\
Top $p$ & 1.0 \\
LR & $1\times 10^{-6}$ \\
KL Coefficient & 0.005 \\
\bottomrule
\end{tabular}
\end{center}
\caption{Basic training hyper-parameters of both GRPO and \method.}
\label{tab: basic hyper-parameters}
\end{table}

\section{Performance on Various Models}
\label{apex: qwen2.5_result}
In addition to Qwen3-4B-Base, Qwen3-8B-Base, and Llama3.2-3B-Instruct, we further evaluate \method on Qwen2.5-3B and Qwen2.5-7B. The results are summarized in Table~\ref{tab:qwen2.5}. Overall, \method consistently achieves the strongest performance across all math reasoning benchmarks.
\setlength\tabcolsep{3.5pt}
\begin{table*}[!ht]
    \centering
    {
    \begin{tabularx}{\textwidth}{l YYYYYYYY}
        \toprule
        \multicolumn{1}{c}{\multirow{2}{*}{\textbf{Model}}} & \multicolumn{4}{c}{\textbf{Pass@1}} & \multicolumn{3}{c}{\textbf{Average@16}} & \multirow{2}{*}{\textbf{Avg.}} \\
        \cmidrule(lr){2-5} \cmidrule(lr){6-8}
        & \textbf{MATH} & \textbf{Minerva} & \textbf{Olympiad} & \textbf{GAO} & \textbf{AMC23} & \textbf{AIME24} & \textbf{AIME25} & \\
        \midrule
        \textbf{Qwen2.5-3B} & 56.00 & 26.10 & 25.33 & 44.16 & 12.97& 1.04& 0.21&23.69 \\
        + GRPO & 65.00 & 27.21 & 27.26 & \underline{56.62} &31.09&4.48 & 1.88&30.51 \\
        + DAPO & \underline{67.40} & 27.21 & 29.48 & \underline{57.40} & 34.06& 4.58&1.04 & 31.60\\
        + GRESO & 65.40 & \textbf{28.31} & \textbf{32.00} & \textbf{58.96} & \underline{37.19}& \underline{6.46}&\underline{2.50} &\underline{32.97} \\
        + GRPO$_{\text{extended}}$ & 66.20 & 27.21 & 26.81 & 56.10 & 27.81&3.96 &\underline{2.29} & 30.05\\
        \rowcolor{gray!10}+ \method (Ours) & \textbf{68.20} & \underline{27.94} & \underline{29.78} & \textbf{58.96} & \textbf{38.91} &\textbf{7.08} &\textbf{3.12} &\textbf{33.43} \\
        \midrule
        \textbf{Qwen2.5-7B} & 62.60 & 27.21 & 29.48 & 51.43 & 19.69& 3.12& 0.42& 27.71\\
        + GRPO & 76.60 & 35.29 & \underline{39.85} & 65.45 & 51.09& 11.46& 5.83& 40.79\\
        + DAPO & 77.00 & 36.40 & 39.41 & 67.01 &53.91 &11.04 & 8.96& 41.96\\
        + GRESO & 77.80 & 37.50 & 40.89 & 66.49 & 48.28& 14.17&6.67 & 41.69\\
        + GRPO$_{\text{extended}}$ & \textbf{78.00} & \underline{36.76} & 38.96& \textbf{69.61} & 51.41& \textbf{15.62}&\underline{5.62}&\underline{42.28} \\
        \rowcolor{gray!10}+ \method & \underline{77.40}& \textbf{37.87}& \textbf{40.89}&\underline{69.39} &\textbf{56.41} &\underline{14.58} & \textbf{10.62}& \textbf{43.88}\\
        \bottomrule
    \end{tabularx}}
    \caption{\textbf{Detailed evaluation results on seven math reasoning benchmarks.} The best and second best results are in \textbf{bold} and \underline{underlined}. (*) Even in the unfavorable setting where DAPO and GRESO are trained for $2\times$ more epochs and GRPO ($n=16$) uses more rollouts, \method still outperforms them on the majority of benchmarks.}
    \label{tab:qwen2.5}
\end{table*}

\section{Success Rate Sensitivity Analysis}
\label{apex: tau} 
To investigate the impact of the success rate threshold $\tau$ on the stability of \method, we conducted experiments with $\tau \in \{0.125, 0.25, 0.375, 0.5\}$.

\noindent\textbf{Results.} The main results are presented in Table~\ref{tab:ablation}. On one hand, $\tau = 0.125$ concentrates the training signal on the most difficult samples near the model's capability frontier, which particularly benefits high-difficulty benchmarks such as Minerva and AMC23. On the other hand, $\tau = 0.5$ provides a broader coverage of both medium- and high-difficulty samples, leading to the best aggregate performance and enhanced stability. Accordingly, we select $\tau = 0.5$ as the default setting for all experiments.

\setlength\tabcolsep{3.5pt}
\begin{table*}[!ht]
    \centering
    {
    \begin{tabularx}{\textwidth}{l YYYYYYYY}
        \toprule
        \multicolumn{1}{c}{\multirow{2}{*}{\textbf{Value}}} & \multicolumn{4}{c}{\textbf{Pass@1}} & \multicolumn{3}{c}{\textbf{Average@16}} & \multirow{2}{*}{\textbf{Avg.}} \\
        \cmidrule(lr){2-5} \cmidrule(lr){6-8}
        & \textbf{MATH} & \textbf{Minerva} & \textbf{Olympiad} & \textbf{GAO} & \textbf{AMC23} & \textbf{AIME24} & \textbf{AIME25} & \\
        \midrule
        $\tau=0.125$ & 67.00 & \textbf{29.78} & 29.04 & 56.88 &\textbf{41.41} &6.67 & 2.50&\underline{33.33} \\
        $\tau=0.250$ &67.40  & \underline{28.31} & \textbf{29.78} & \underline{57.92} & 36.88&\textbf{7.50} & 2.50& 32.90\\
        $\tau=0.375$ & \underline{68.00} & 26.47 & \textbf{29.78} & 56.36 & 37.19& 6.04& \underline{2.92}&32.39 \\
        $\tau=0.500$ & \textbf{68.20} & 27.94 & \textbf{29.78} & \textbf{58.96} &\underline{38.91} & \underline{7.08}& \textbf{3.12}& \textbf{33.43}\\
        \bottomrule
    \end{tabularx}}
    \caption{\textbf{Validation result under different $\tau$}.}
    \label{tab:ablation}
\end{table*}

\section{Computational Overhead}
\label{anaysis:computation}
We report detailed runtime and memory analyses of \method compared with standard GRPO under identical hardware configurations, utilizing the \texttt{verl} framework on $8\times$ NVIDIA A800 GPUs. \method incurs a higher wall-clock cost per update to prioritize signal quality, the specific average step times are detailed in Table~\ref{tab:efficiency_comparison}.

\begin{table*}[!ht]
\centering
\begin{tabular}{llccc} 
\toprule
\textbf{Model} & \textbf{Algorithm} & \textbf{Group Size ($G$)}  & \textbf{Avg. Step Time (s)} & \textbf{Relative Cost} \\
\midrule
\multirow{3}{*}{Qwen2.5-3B} 
    & GRPO & 8 & 214 & $1.0\times$ \\
    & GRPO$_{\text{extended}}$ & 16 & 384 & $1.79\times$ \\
    & \method (Ours) & 8 & 333 & $1.56\times$ \\
\midrule
\multirow{3}{*}{Qwen2.5-7B} 
    & GRPO & 8 & 344 & $1.0\times$ \\
    & GRPO$_{\text{extended}}$ & 16 & 523 & $1.52\times$ \\
    & \method (Ours) & 8 & 463 & $1.35\times$ \\
\midrule
\multirow{3}{*}{Qwen3-4B-Base}    
    & GRPO & 8 & 331 & $1.0\times$ \\
    & GRPO$_{\text{extended}}$ & 16 & 524 & $1.58\times$ \\
    & \method (Ours) & 8 & 475 & $1.44\times$ \\
\midrule
\multirow{3}{*}{Qwen3-8B-Base}    
    & GRPO & 8 & 477 & $1.0\times$ \\
    & GRPO$_{\text{extended}}$ & 16 & 802 & $1.79\times$ \\
    & \method (Ours) & 8 & 726 & $1.62\times$ \\
\midrule
\multirow{3}{*}{Llama-3.2-3B-Instruct}    
    & GRPO & 8 & 229 & $1.0\times$ \\
    & GRPO$_{\text{extended}}$ & 16 & 338 & $1.48\times$ \\
    & \method (Ours) & 8 & 291 & $1.27\times$ \\
\bottomrule
\end{tabular}
\caption{\textbf{Computational efficiency comparison on 8 $\times$ A800.} $G$ denotes the group size.}
    \vspace{-0.3cm}
\label{tab:efficiency_comparison}
\end{table*}

\setlength\tabcolsep{3.5pt}
\begin{table*}[!ht]
    \centering
    {
    \begin{tabularx}{\textwidth}{l YYYYYYYY}
        \toprule
        \multicolumn{1}{c}{\multirow{2}{*}{\textbf{Method}}} & \multicolumn{4}{c}{\textbf{Pass@1}} & \multicolumn{3}{c}{\textbf{Average@16}} & \multirow{2}{*}{\textbf{Avg.}} \\
        \cmidrule(lr){2-5} \cmidrule(lr){6-8}
        & \textbf{MATH} & \textbf{Minerva} & \textbf{Olympiad} & \textbf{GAO} & \textbf{AMC23} & \textbf{AIME24} & \textbf{AIME25} & \\
        \midrule
        GRPO & 65.00 & 27.21 & 27.26 & 56.62 &31.09 &4.48&1.88&30.51 \\
        Resampling & 66.40 & 26.47 & 29.48 & 57.92 & 37.50&6.67&2.50&32.42 \\
        Random Pruning & 65.80 & 26.10 & 29.19 & 57.66 & 37.50&5.83&2.92& 32.14\\
        Gradient-based Pruning & 65.60 & 26.84 & 29.78 & 58.18 & 37.81&6.67&1.88&32.39 \\
        \method & \textbf{68.20} & \textbf{27.94} & \textbf{29.78} & \textbf{58.96} &\textbf{38.91} & \textbf{7.08}& \textbf{3.12}& \textbf{33.43}\\
        \bottomrule
    \end{tabularx}}
    \caption{\textbf{Validation result under different pruning strategies.}}
    \vspace{-0.6cm}
    \label{tab:pruning_strategies}
\end{table*}

\noindent\textbf{Results.} \method incurs a computational overhead ranging from $1.27\times$ to $1.62\times$ compared to standard GRPO ($G=8$). 
Notably, this remains significantly more efficient than the brute-force approach of doubling sample size ($G=16$). 
Crucially, this additional computation translates directly into enhanced reasoning capabilities, evidenced by the 2--3\% absolute accuracy gains reported in Table~\ref{tab:main}. 
These results demonstrate that \method offers a superior trade-off between efficiency and effectiveness, successfully converting marginal temporal costs into substantial reasoning capabilities.

\section{Pruning Strategies Analysis}
\label{anaysis:masking}
To demonstrate the effectiveness of our pruning strategy, we compare \method with three representative alternatives:
(1) \textbf{Resampling}, which uses an additional $n$ rollouts to replace unsuccessful rollouts with successful ones;
(2) \textbf{Random Pruning}, which prunes the same fraction of tokens uniformly at random per instance; and
(3) \textbf{Gradient-based Pruning}, which prunes tokens with the smallest gradient norm.
For fairness, all methods share the same training setup and pruning ratio, and we evaluate them on seven benchmarks.

\begin{table}[t!]
    \centering
    \begin{tabular}{lccc}
        \toprule
        \textbf{Metric} & \textbf{Qwen3-4B} & \textbf{Qwen3-8B} & \textbf{Llama3.1-8B} \\
        \midrule
        F1 & 0.9875 & 0.9871 & 0.9877 \\
        Precision & 0.9880 & 0.9875 & 0.9883 \\
        Recall & 0.9871 & 0.9866 & 0.9871 \\
        \bottomrule
    \end{tabular}
    \caption{BERTScore between original and pruned prompts exceeds 0.98, showing semantic preservation.}
    \label{tab:bertscore}
    \vspace{-0.6cm}
\end{table}

\noindent\textbf{Results.} We present the results in Table~\ref{tab:pruning_strategies}. Our observations are as follows: \method consistently outperforms the three baselines, achieving the best accuracy on seven benchmarks, which indicates more consistent and stable improvements than competing methods.

\section{Semantic Preservation of Interference Purification}
\label{app:semantic_preservation}

A natural concern is whether removing interference tokens might distort 
prompt semantics. We provide two pieces of evidence showing that \method 
preserves semantic integrity.

\paragraph{Functional Equivalence.} 
In RLVR, rewards are computed against the unique ground-truth answer. 
If pruning distorted critical semantics, the generated reasoning would 
diverge from the correct solution and yield zero reward. The substantial 
post-pruning accuracy gains (Figure~\ref{fig:main}) 
therefore serve as functional proof that pruned prompts remain 
semantically equivalent to the originals.

\paragraph{Quantitative Verification.} 
We compute BERTScore between original and pruned prompts across three 
backbones (Table~\ref{tab:bertscore}). All F1 scores exceed 0.98, 
confirming strong semantic equivalence. Qualitatively, removed tokens 
are predominantly formatting noise (e.g., \texttt{\textbackslash\textbackslash}, 
\texttt{\$\$}), template redundancy (e.g., \texttt{Let}, \texttt{Given}), 
and irrelevant entities rather than semantically load-bearing content.

\section{Checklist}
\paragraph{Potential Risks} Our work does not involve any identifiable ethical or legal risks.
\paragraph{Artifacts} We check that the data does not contain any information that names or uniquely identifies individual people or offensive content. All models and datasets used in this work comply
with their respective open-source or research licenses. We ensure that all artifacts are used strictly
within the permitted scope of their terms. The Code
we released will be under a permissive open-source 
license, enabling reproducibility and reuse. Documentation for all artifacts will be updated and made available in the project’s GitHub repository upon release.
\paragraph{AI Assistants}
We used AI assistants (ChatGPT) solely for textual
and grammatical refinement, without influencing
the core content or experiments.

\end{document}